# Fault Diagnosis of Helical Gear Box using Large Margin K-Nearest Neighbors Classifier using Sound Signals

M. Amarnath[1], S. Arunav[2], *Hemantha Kumar[3], V. Sugumaran[4]
and G.S Raghvendra[5]

[1]Indian Institute of Information Technology Design and Manufacturing Jabalpur, Jabalpur, Madyapradesh, India,

[2, 5]Birla Institute of Technology and Science – Pilani, K K Birla Goa Campus, Goa – 403726, India.

[3]National Institute of Technology Karnataka, Surathkal, Mangalore - 575025, Karnataka, INDIA. Email: hemanta76@gmail.com
(* Corresponding author)

[4]VIT University, Chennai Campus, Vandalur-kelambakam road, Chennai-600048, v_sugu@yahoo.com

## Abstract

Gear drives are one of the most widely used transmission system in many machinery. Sound signals of a rotating machine contain the dynamic information about its health conditions. Not much information available in the literature reporting suitability of sound signals for fault diagnosis applications. Maximum numbers of literature are based on FFT (Fast Fourier Transform) analysis and have its own limitations with non-stationary signals like the ones from gears. In this paper, attempt has been made in using sound signals acquired from gears in good and simulated faulty conditions for the purpose of fault diagnosis through a machine learning approach. The descriptive statistical features were extracted from the acquired sound signals and the predominant features were selected using J48 decision tree technique. The selected features were then used for classification using Large Margin K-nearest neighbor approach. The paper also discusses the effect of various parameters on classification accuracy.

## 1. Introduction

Helical gear box condition monitoring has received considerable attention for many years. Gears are the most important and frequently encountered components in the vast majority of rotating machines. Their load carrying capacity and reliability is of prime importance for the overall machine performance. Hence, fault diagnosis of such machine elements has been the subject matter of extensive research (Drosjack and Houser [9]).Gear failures can be caused by several factors such as incorrect design or installation, acid corrosion, poor lubrication etc. Vibration and sound monitoring has been widely reported as being a useful technique for the diagnosis of the condition of rotating machines. It can help fault detection before significant damage occurs. More efficient maintenance scheduling can be planned if accurate information about a machine's condition is known and if online monitoring is used .The traditional pattern recognition includes a large collection of very different types of mathematical tools (preprocessing, extraction of features and final recognition). In many cases it is difficult to say what kind of tool would be the best for a particular problem Fault classification techniques have been used in a wide range of pattern recognition applications including sound vibration monitoring. The contributions of some authors (Bo et al. [5], Samanta [14], Chen and Wang [7] and Paya et al. [10]) reveal the application of neural networks to online condition monitoring of rotating machinery to have very high success rates. ANNs consequently appear to be a possible solution to gear diagnostics problem as they could allow real-time online condition monitoring at a reasonably low cost (Yang et al. [22]). Paya et al. [10] carried out investigations to study both bearing and gear faults introduced separately as a single fault and then together as multiple faults in the drive line. The real time signals obtained from the driveline were preprocessed by wavelet transforms for the neural network to perform fault detection and identification of the exact kind of fault occurring in the model drive line. The authors summarized the results of their research for distinguishing between different kinds of faults viz., good gear, blip gear, shaved gear and one with inner race defect. An overall success rate of 96% was achieved on test by back propagation network which gave the details of exact kind of fault in the driveline. Baydar and Ball [4] demonstrated the results of fault diagnosis experiments conducted on two stage helical gearbox. Authors have considered sound and vibration signals to detect local faults in helical gear tooth. Sound and vibration signals acquired from the gearbox were processed using Morlet wavelet. Amplitude and phase maps obtained from wavelet analysis provided a good visual inspection tool to detect faults in the early stage. Vyas and Satishkumar [19] carried out experiments to automate the fault detection procedure in rotating machinery. A back propagation learning algorithm and a multilayer network were employed for fault detection. Five different types of faults were introduced in the experimental setup and five statistical moments of vibration signals were employed to train the network. An overall success rate of 90% was obtained in this work. Chen and Wang [7] dealt with multi layer perceptron (MLP) pattern classifiers for wavelet map interpretation and their application as a tool for mechanical fault detection. Features for neural networks were extracted from instantaneous scale distribution. This study was undertaken to simplify the difficulties in inspecting complicated wavelet patterns in time-scale domain. The authors highlighted the details of construction, training and testing multilayer perceptron based classifiers for diagnosis of gear faults. Ramroop *et al.* [13] conducted experimental investigation to detect faults in multistage industrial gearbox, sound signals were acquired from the gearbox under near field condition, and fast Fourier transform (FFT) method used extract fault related features from these signals. This paper provided a series of best practice guidelines for implementation sound condition monitoring technique to detect local faults in industrial gearbox. Wuxing *et al*. [21] conducted experiments on a gearbox to classify the gear faults using cumulants and the radial basis function (RBF) network. The cumulants were calculated from the vibration signals collected from the inspected gearbox and were used as input features to an ANN. The radial basis function network was then used as a classifier for various operating conditions of the helical gear box. e.g., normal, spalling, one worn tooth condition and two worn teeth condition. The authors concluded that the method of fault classification by combining cumulants and the radial basis function network is promising and achieves better accuracy than many of the current methods available. Samanta [23] presented an experimental study to compare the performance of gear fault detection and classification using ANN and SVM. The time domain vibration signals of a rotating machine with normal and defective gears were pre-processed for feature extraction. The role of different vibration signals at normal and light loads were investigated in this work. SVM shows better classification accuracy than ANN. In addition, genetic algorithms (GA) were used to improve accuracy of fault classification. With GA based selection, the performance of ANN and SVM showed comparatively equal accuracy in results. Yang *et al.* [23] presented a novel scheme to detect faults in reciprocating compressors of refrigerators. The vibration and noise signals were wavelet transformed to find diagnostic information. Further the statistical features of wavelets were used for fault classification using ANN and SVM techniques. A high accuracy in classification of faults was obtained using SVM technique. Shin *et al.* [15] adopted SVM technique for detection and classification of faults in electro mechanical machinery using vibration parameters. Multilayer perceptrons of ANN technique was also included in the diagnosis program. The results concluded that the classification of faults using SVM was superior to that of MLP of ANN techniques. Sugumaran *et al.* [17] employed proximal support vector machines (PSVM) and SVM to classify faults in bearings. The authors compared the results of PSVM and SVM. PSVM was found to have less iterations and faster learning as compared to SVMs in fault classification. A novel method to diagnose faults in rotating machinery was proposed by Qiao *et al.* [11]. Improved wavelet package transform was used in to extract the salient frequency band features from the vibration signals. SVM ensemble technique was adopted in fault classification, which provides promising results in diagnosis of machinery. Amarnath and Praveen Krishna [3] carried out experiments to detect faults in ball bearing and gears using sound signals. Emperical mode decomposition (EMD) method was used to detect local faults in bearings and gears, EMD based statistical parameters such as kurtosis, root mean square, skewness, crest factor and impulse factor values were extracted from sound signals, these fault related features showed comparatively better fault diagnostic information than that of statistical parameter values of unprocessed sound signals. Amarnath *et al.* [2] used acoustic signals acquired from near field area of bearings in good and simulated faulty conditions for the purpose of fault diagnosis through machine learning approach. Sugumaran *et al.* [18] used vibration signals acquired from gears in good and simulated faulty conditions for the purpose of fault diagnosis through J48 decision tree algorithm.

In the present study, an attempt is made to exploit sound signals for the purpose of fault diagnosis of helical gear box. To extract some meaningful features, descriptive statistical features like mean, median, kurtosis etc., were used. Important features were selected and classification was carried out using the novel Large Margin K-nearest neighbor algorithm with varying number of neighbors and size of training set using random sub sampling. A modified version of the pseudo code given by Bremer *et al* [6] along with modifications proposed by Cost and Salzberg [8] is used in this study in order to train and test the classifier. The results for both vibration and sound signals are plotted as a function of test case size versus classification accuracy percentage.

## 2. Experimental Setup

Fig. 1 shows the experimental set up (Amarnath *et al.* [1]). It consists of a 5 HP two stage helical gearbox driven by a 5.5 HP 3-phase induction motor at 1200 rpm. The mechanical output of the gearbox is used to drive a D.C generator and the output power of the generator is dissipated in a resistor bank, which provides torque load to the generator and gearbox. This arrangement does not provide any additional vibration to the test rig. The motor, gearbox and generator are mounted on stiffened I-beams, which are anchored to a massive concrete block. An accelerometer B&K 4332 is stud-mounted to measure the vertical vibration signals generated on the bearing housing of the 16 teeth pinion. Meshing gear frequencies are calculated at 320 Hz and multiples. Different data sets were collected when the helical gear train was working at normal, 10%, 20%, 40%, 60%, 80% and 100% tooth removal conditions (Badyar and Ball [4]). A total of 30 data sets were collected for each operating condition. The signals were truncated to 3 kHz using a low pass filter and sampled at 8 kHz. The accelerometer outputs were conditioned using B&K TYPE 2626 charge amplifier.

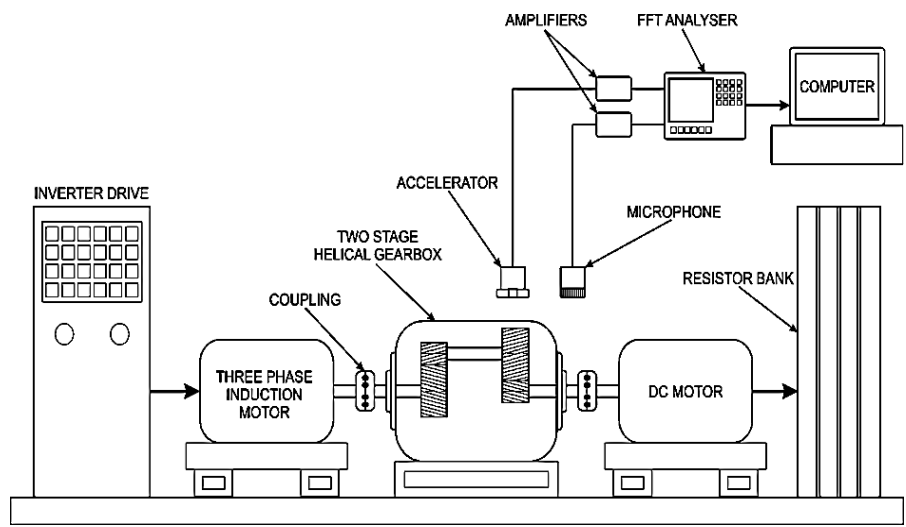

Fig.1 Experimental setup of two stage helical gearbox

Table 1 Specifications of helical gearbox

|  | First stage | Second stage |
|---|---|---|
| Number of teeth | 44/13 | 73/16 |
| Pitch circle diameter (mm) | 198 /65 | 202 /48 |
| Pressure angle (°) | 20 | 20 |
| Helix angle | 20 | 15 |
| Modules | 4.5/ 5 | 2.75 / 3 |
| Speed of shafts | 80 rpm (input) | 1200 rpm (output) |
| Mesh frequency | 59 Hz | 320 Hz |
| Step - up ratio | 1:15 | |
| Rated power | 5 HP | |
| Power Transmitted | 2.6 HP | |

The pinion is connected to a D.C motor (which is used as generator) to generate 2 kW power, which is dissipated in a resistor bank. Hence, the actual load on the gearbox is only 2.6 HP which is 52% of its rated power 5 HP. In industrial environment utilization of load varies from 50% to 100%. In the case of traditional dynamometer, additional torsional vibrations can occur due to torque fluctuations. This is avoided in this case by using D.C motor and resistor bank. Tyre couplings are fitted between the electrical machines and gear box so that backlash in the system can be restricted to the gears. The motor, gear box and generator are mounted on I-beams, which are anchored to a massive foundation. Vibration signals are measured using a Bruel and Kjær accelerometer which is installed close to the test bearing. Signals are sampled at a sampling frequency of 8.2 kHz. The experimental setup with equipment and sensors is shown in Fig. 2.

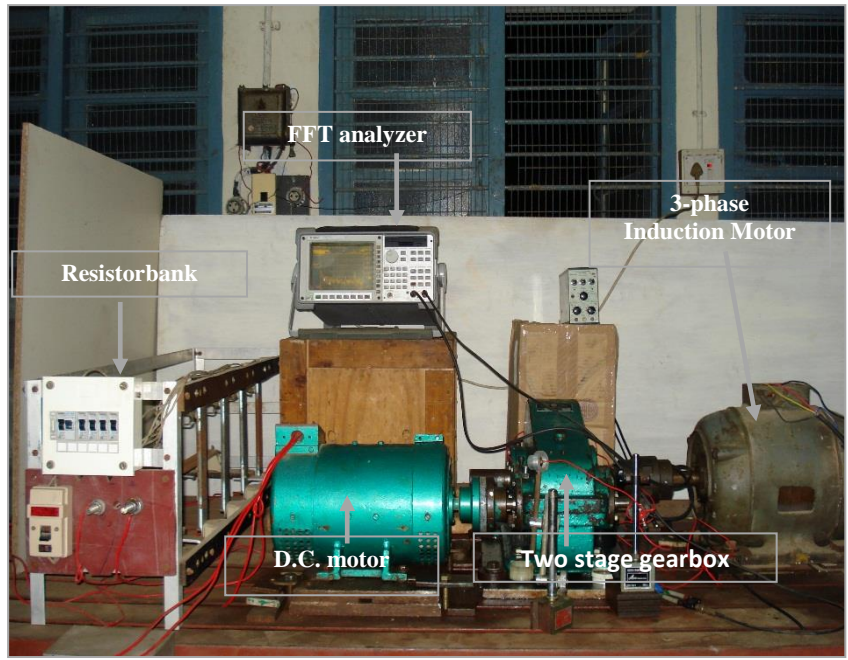

Fig. 2 Photograph of experimental set up with sensors and equipments

Overhaul time of a new gear box is more than one year. It is very difficult to study the fault detection procedures without seeded fault trials. Local faults in a gear box can be classified into three categories. (a) Surface wear spalling (b) cracked tooth and (c) loss of a part of tooth due to breakage of tooth at root or at a point on working tip (broken tooth or chipped tooth). There are different methods to simulate faults in gears viz. electric discharge machining (EDM), grinding, adding iron particles in gearbox lubricant and over loading the gear box i.e. accelerated test condition. The simplest approach is partial tooth removal. This simulates the partial tooth break, which is common in many industrial applications (Staszewski *et al.* [16] and Yesilyurt *et al.* [24])).

## 3. Statistical Feature Extraction

From the vibration signals, descriptive statistical parameters such as mean, median, mode, kurtosis, skewness, standard error, standard deviation, minimum, maximum, sum, and range are computed to serve as features. They are named as 'statistical features' here. Brief descriptions about the extracted features are given below.

(a) **Standard error**: Standard error is a measure of the amount of error in the prediction of *y* for an individual *x* in the regression, where *x* and *y* are the sample means and '*n*' is the sample size.

$$Standard\ error = \sqrt{\frac{1}{n-2}\sum\left((y-\bar{y})^2 - \frac{\sum[(x-\bar{x})(y-\bar{y})]^2}{(x-\bar{x})^2}\right)} \quad (1)$$

(b) **Standard deviation**: This is a measure of the effective energy or power content of the vibration signal. The following formula was used for computation of standard deviation.

$$Standard\ deviation = \sqrt{\frac{\sum x^2 - (\sum x)^2}{n(n-1)}} \quad (2)$$

(c) **Sample variance**: It is variance of the signal points and the following formula was used for computation of sample variance.

$$Sample\ Variance = \frac{\sum x^2 - (\sum x)^2}{n(n-1)} \quad (3)$$

(d) **Kurtosis**: Kurtosis indicates the flatness or the spikiness of the signal. Its value is very low for normal condition of the bearing and high for faulty condition of the bearing due to the spiky nature of the signal.

$$Kurtosis = \left(\frac{n(n+1)}{(n-1)(n-2)(n-3)}\sum\left(\frac{x_i-\bar{x}}{s}\right)^4\right) - \frac{3(n-1)^2}{(n-2)(n-3)} \quad (4)$$

where '*s*' is the sample standard deviation.

(e) **Skewness**: Skewness characterizes the degree of asymmetry of a distribution around its mean. The following formula was used for computation of skewness.

$$Skewness = \frac{n}{n-1}\sum\left(\frac{x_i-\bar{x}}{s}\right)^3 \quad (5)$$

(f) **Range**: It refers to the difference in maximum and minimum signal point values for a given signal.

(g) **Minimum value**: It refers to the minimum signal point value in a given signal. As the bearing parts (inner race, outer race) get degraded, the vibration levels seem to go high. Therefore, it can be used to detect faulty bearing condition.

(h) **Maximum value**: It refers to the maximum signal point value in a given signal.

(i) **Sum**: It is the sum of all feature values for each sample.

## 4. Large Margin K-Nearest Neighbors Algorithm

The large Margin K-nearest neighbor algorithm is a variant of the original K-nearest neighbour algorithm for classification [Bremer et al [6]]. The basis of the algorithm lies in the construction of a distance metric. Given any two objects (vectors in a feature space) x and y a metric is defined as *d(x, y)* provided it satisfies the following criteria:-

1. $d(x, y) \geq 0$   (*non-negativity*, or separation axiom)

2. $d(x, y) = 0$  if and only if  $x = y$   (coincidence axiom)

3. $d(x, y) = d(y, x)$   (*symmetry*)

4. $d(x, z) \leq d(x, y) + d(y, z)$   (*Sub-additivity / triangle inequality*).

A distance measure which satisfies conditions 1, 2 and 4 is called a pseudo metric. In statistical data mining algorithms a metric or a pseudo metric can be used as a quantitative dissimilarity measure.

Let each object (observation) be defined as a vector in the feature space of its dimensions. Also let an integer value k (the number of nearest neighbors to be considered) be defined. Also, let the class variable take n possible values denoted by $C_i$ where $i \in \{1, n\}$

## 4.1 Training Phase

In the training phase a certain subset of vectors are chosen from the original data set as the training set. Let this subset be T. Now the first step of the algorithm requires the construction of a pseudo metric. The metric can be written as:-

$$d(\vec{x}_i, \vec{x}_j) = (\vec{x}_i - \vec{x}_j) M (\vec{x}_i - \vec{x}_j)^T \tag{6}$$

Here $\vec{x}_i$ and $\vec{x}_i$ are two observations or instances. In the special case of M being identity, the metric is same as Euclidean distance. The algorithm makes a distinction between two types of instances called target neighbors and impostors which are described in the next two subsections.

### *4.1.1 Target Neighbors*

Target neighbors are instances which are selected before learning. Each instance $\vec{x}_i$ has exactly k different target neighbors within D, which all shares the same class label $y_i$. The target neighbors are the data points that should become nearest neighbors under the learned metric. Let us denote the set of target neighbors for a data point $\vec{x}_i$ as $N_i$.

### *4.1.2 Impostors*

An impostor of a data point $\vec{x}_i$ is another data point $\vec{x}_j$ with a different class label (i.e $y_i \neq y_j$) where $\vec{x}_j$ is one of the k nearest neighbors of $\vec{x}_i$. During learning the algorithm tries to minimize the number of impostors for all data instances in the training set.

The Large Margin Nearest Neighbor algorithm optimizes the matrix M using semi-definite programming. The objective has two components: For every data observation $\vec{x}_i$, the target neighbors should be as close as possible and the impostors should be as far away as possible simultaneously. The learned pseudo metric causes the input vector $\vec{x}_i$ to be surrounded by training instances of the same class (i.e target neighbors). If a test point is being classified, it would be classified correctly under the new metric distance scheme, since it will be surrounded by its target neighbors.

### 4.2 Problem Restatement and Solver Algorithm

The problem can be restated in terms of an optimization problem. The first optimization goal is achieved by minimizing the average distance between instances and their target neighbors

$$minimize \ \sum_{i,j \ \in N_i} d(\vec{x}_i, \vec{x}_j) \tag{7}$$

The second goal is achieved by constraining impostors $\vec{x}_l$ to be one unit further away than target neighbors $\vec{x}_j$ (and therefore pushing them out of the local neighborhood of $\vec{x}_i$). The resulting inequality constraint can be stated as:

$$\forall_{i,j \in N \ l, y_l \neq y_i} \ d(\vec{x}_i, \vec{x}_j) + 1 \leq d(\vec{x}_i \vec{x}_l) \tag{8}$$

The margin of exactly one unit fixes the scale of the matrix M. Any alternative choice $c > 0$ would result in a rescaling of M by a factor of $1/c$.

The final optimization problem thus problem becomes:

$$minimize \ \sum_{i,j \ \in N_i} d(\vec{x}_i, \vec{x}_j) + \sum_{i,j,l} \varepsilon_{i,j,l} \tag{9}$$

Subject to:-

$$\forall_{i,j \in N \ l, y_l \neq y_i} \ d(\vec{x}_i, \vec{x}_j) + 1 \leq d(\vec{x}_i \vec{x}_l) + \varepsilon_{i,j,l} \tag{10}$$

$$\varepsilon_{i,j,l} \geq 0 \tag{11}$$

$$M \geq 0 \tag{12}$$

Here the slack variables $\varepsilon_{i,j,l}$ absorb the amount of violations of the impostor constraints (i.e the slack variables minimize the errors caused by the impostor instances). The overall sum of distance and impostor errors is minimized. Constraint (12) ensures that M is positive semi-definite. This optimization problem is an instance of semi - definite programming (SDP). Although SDPs tend to suffer from high computational complexity, this particular SDP instance can be solved very efficiently due to the underlying geometric properties of the problem. In the present case a particularly efficient gradient based solver proposed by Weinberger et al. [20] is used. Certain modifications are done to the algorithm in order to guarantee termination and save computation time and resources.

We need to find a matrix M that minimizes the distance measure as defined in equation (6). We will call this M matrix the optimal matrix .The solver algorithm is iterative and is based on the gradient descent approach. A pseudo code description of the algorithm is given in appendix A along with detailed explanations of the data structures used. The algorithm is governed by the following equations:-

Firstly the error introduced by impostor instances can be written in the form of equation (13),

$$\varepsilon_{i,j,l}(M) = maximum\left(0, 1 + d_M^2(\vec{x}_i, \vec{x}_j) - d_M^2(\vec{x}_i, \vec{x}_l)\right) \tag{13}$$

Here M is the optimal matrix as required in equation 6. Also:-

$$d_M^2(\vec{x}_i, \vec{x}_j) = (\vec{x}_i - \vec{x}_j)M(\vec{x}_i - \vec{x}_j)^T = trace(C_{ij}M) \tag{14}$$

$$C_{ij} = (\vec{x}_i - \vec{x}_j)(\vec{x}_i - \vec{x}_j)^T \tag{15}$$

Equation (14) is just a paraphrasing of equation (6) using trace operation on matrices. Now consider μ as the rate of gradient descent. The objective function can now be paraphrased as:-

$$\varepsilon(M) = \sum_{j \to i}\left(d_M^2(\vec{x}_i, \vec{x}_j) + \mu(1 - y_{il})\varepsilon_{i,j,l}(M)\right) \tag{16}$$

Here $y_{il} \in \{0,1\}$ and $y_{il} = 1$ if $\vec{x}_i$ and $\vec{x}_l$ have the same class label and 0 otherwise. Also $j \to i$ indicates that $\vec{x}_j$ is a target neighbor of $\vec{x}_i$ .Moreover the constraint (8) changes to

$$d_M^2(\vec{x}_i, \vec{x}_j) - d_M^2(\vec{x}_i, \vec{x}_l) \geq 1 - \varepsilon_{i,j,l} \tag{17}$$

Whereas constraints (9) and (10) remain the same.

Now the gradient matrix is calculated using the following equation:-

$$G_t = \sum_{j \to i} C_{ij} + \mu \sum_{(i,j,l) \in N_t} (C_{ij} - C_{il}) \qquad (18)$$

The 3 – tuple $(i, j, l) \in N_t$ if $\varepsilon_{i,j,l}(M^t) \geq 0$ and $M^t$ is the matrix M at the $t^{th}$ iteration.

Finally the Matrix M is recalculated at every iteration until a user defined convergence criteria is met:-

$$M_{t+1} = M_t + P_s(M_t - \propto G_{t+1}) \qquad (19)$$

Here $\propto$ is a user defined parameter and $P_s$ is the projection of M to all semi-definite cones. The projection to all semi definite cones arises from the following equations,

$$P_S(M^t) = V\Delta^+V^T \qquad (20)$$

$$M^t = V\Delta V^T \qquad (21)$$

Also $\Delta = \Delta^+ + \Delta^-$ and $\Delta^+ = \max(0, \Delta)$. Here $\Delta$ is a diagonal matrix of all the eigenvalues of the optimal matrix. $\Delta^+$ is the diagonal matrix with all the negative entries of $\Delta$ replaced with 0. In other words it is the diagonal matrix of positive eigenvalues of the optimal matrix. $V$ is the square matrix of all the eigenvectors of the optimal matrix stacked together column wise.

Once the pseudo metric is ascertained we move on to the Test phase.

### 4.3 Test Phase

For $\forall x \in T^C$ do:-

1. Determine the weighted k-nearest neighbors (k elements of the training set with least pseudo metric distance). One can do this by repeatedly picking the element with the minimum distance (closest neighbor) and repeat it k times without replacement. Moreover while picking the minimum distance incorporate the following weight measure (Cost and Salzberg [8]),

$$W(C_i) = W(C_i) + 1/d(\vec{x}_i, \vec{x}_j) \tag{22}$$

Here $\vec{x}_i$ is the test instance and $\vec{x}_j$ is the training instance which is considered as one of the k-nearest neighbors.

2. Determine the weight of instances of each $C_i$ denoted by $W(C_i)$. Now let $C_i =$ argmax $W(C_i)$ $\forall i \in \{1, n\}$. The previous sentence means - choose by majority voting which class has highest representation in the k-nearest neighbors and classify the test example by that.

As stated previously, a pseudo code description of the algorithm is given in Appendix (A) along with modifications made for this study.

## 5. Results and Discussion

The sound signals were recorded for normal and abnormal conditions of helical gear box. Totally 420 samples were collected; out of which 60 samples were from Healthy condition. For faulty load with 10%, 20%, 40%, 60%, 80% and 100% fault, 60 samples from each condition were collected. The statistical features were treated as features (attributes) and act as inputs to the algorithm. The corresponding status or condition (10% fault, 20 % fault, 40% fault, 60% fault, 80% fault, 100% fault and healthy) of the classified data will be the required output of the algorithm. This input and corresponding output together forms the dataset. The dataset is used with decision tree J48 algorithm for generating the decision tree for the purpose of feature selection (Quinlan [12]). Although the nodes closer to the root are more significant, all nodes in the tree are given equal importance for feature subset selection in order to maintain simplicity of the code. Of all the fourteen features extracted, only standard error was discarded.

Once the features were selected, the large margin k- nearest neighbor classifier was used for both training and testing purposes. The number of objects (training set size) for testing was varied from 1 to 59 and found that when it is 50 and when k = 1, the algorithm gives best classification accuracy of 94.3% for sound signals. Out of 420 data points 396 data points were classified successfully.

The following graph is a graph for classification accuracy(in percentage) versus Test set size for a few representative values of nearest neighbors (i.e. k values).

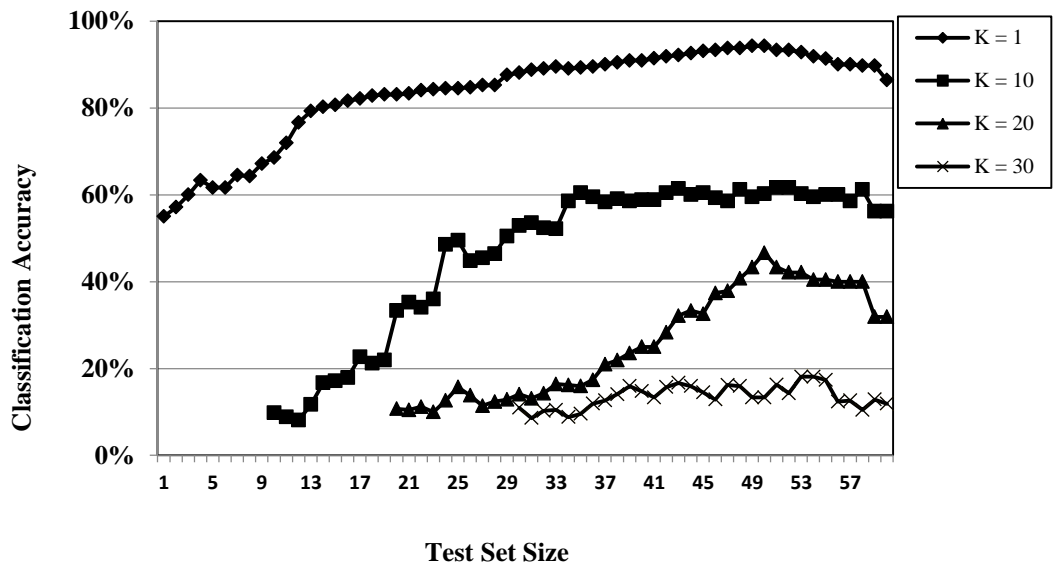

Fig. 3 Test set size versus classification accuracy

The confusion matrix is plotted for k = 1 and test set size = 50. It is a quantitative summary of the classification details.

| Confusion Matrix | | | | | | | |
|---|---|---|---|---|---|---|---|
| **A** | **B** | **C** | **D** | **E** | **F** | **G** | |
| 44 | 3 | 4 | 2 | 2 | 5 | 0 | A = 10 percent fault |
| 1 | 56 | 1 | 0 | 0 | 2 | 0 | B = 20 percent fault |
| 0 | 0 | 58 | 0 | 0 | 2 | 0 | C = 40 percent fault |
| 0 | 0 | 0 | 60 | 0 | 0 | 0 | D = 60 percent fault |
| 0 | 0 | 1 | 0 | 59 | 0 | 0 | E = 80 percent fault |
| 0 | 0 | 0 | 0 | 1 | 59 | 0 | F = 100 percent fault |
| 0 | 0 | 0 | 0 | 0 | 0 | 60 | G = Healthy |

Table 2 – Confusion matrix for sound signals

The interpretation of the confusion matrix is as follows:-

- The diagonal elements in the confusion matrix show the number of correctly classified instances. The rows correspond to the classification achieved by the LMNN algorithm (the predicted classes) and the columns correspond to the actual classes.

- Out of the 420 data instances, 396 have been classified correctly.

## 6. Conclusions

Gears are important machine elements in industrial machinery which is subjected to wear and tear. This paper presented an algorithm based interpretation of vibration signals for automated evaluation of gear condition. From acquired vibration data, a model was built using data modeling technique. Decision tree algorithm was used for feature subset selection and large margin k – nearest neighbor algorithm was used for classification of the condition of the gear. The built model was tested with all possible combinations of nearest neighbor and test set size value and an accuracy of 94.2% was achieved for k = 1 and test set size = 50. Hence, the results of the large margin k - nearest neighbor algorithm can be practically used for diagnosing the condition of the gears successfully.

## 7. References

[1] Amarnath, M., Sujatha, C., and Swarnamani, S., Experimental studies on the effects of reduction in gear tooth stiffness and lubricant film thickness in a spur geared system, *Tribology International*, Vol.42(2), pp. 340-352, 2009.

[2] Amarnath, M., Sugumaran, V., and Hemantha kumar (2013), Exploiting Sound Signals for Fault Diagnosis of Bearings Using Decision Tree, *Measurement*, Vol. 46, pp. 1250 – 1256, 2013.

[3] Amarnath, M., and Praveen, K., Detection and diagnosis of surface wear failure in a spur geared system using EEMD based vibration signal analysis, *Tribology International*, Vol. 61, pp. 224-234, 2013.

[4] Baydar, N., and Ball, A., Detection and diagnosis of gear failure via vibration and acoustic signals using wavelet transform, *Mechanical Systems and Signal Processing,* Vol. 17(4), pp. 787-804, 2003.

[5] Bo, L., Mo, Y., and James C. H., Neural network based motor rolling bearing fault diagnosis, *IEEE Transaction on Industrial Electronics*, Vol. 47 (5), pp. 1060-1069, 2000.

[6] Bremner, D., Demaine, E., Erickson, J., Iacono, J., Langerman, S., Morin, P., and Toussaint, G., Output-sensitive algorithms for computing nearest-neighbor decision boundaries, *Discrete and Computational Geometry,* Vol. 33(4), pp. 593–604, 2005.

[7] Chen, D. and Wang, W. J., Classification of wavelet patterns using multilayer neural networks, *Mechanical Systems and Signal Processing*, Vol. **16**(4), pp. 695-704, 2002.

[8] Cost, S., and Slazberg, S., A weighted Nearest Neighbor Algorithm for learning with symbolic features. *Machine learning*, pages 10:57 − 78, 1993.

[9] Drosjack, M. J. and Houser, D. R., An experimental and theoretical studies of the effect of simulated pitch line pitting on the vibration of a geared system, *ASME publication report 77-DET-123*, 1977.

[10] Paya, B. A., East, I. I., and Badi, M. N. M., Artificial neural network based fault diagnostics of rotating machinery using wavelet transform as a preprocessor, *Mechanical Systems and Signal Processing*, Vol. 11(5), pp. 751-765, 1997.

[11] Qiao, H., Zhengjia, H., Zhousuo, Z., and YanyangZi, Fault diagnosis of rotating machinery based on improved Wavelet package transform and SVMs ensemble, *Mechanical Systems and Signal Processing,* Vol. 21(2), pp. 688-705, 2007.

[12] Quinlan, J. R., Induction of Decision Trees, *Machine Learning*, Vol.1, pp. 81-106, 1986.

[13] Ramroop, G., Liu, K., Gu, F., Paya B. S., and Ball, A. D., Airborne Acoustic Condition Monitoring, *www.maintenenceengineering.com* (Accessed Dec. 2003).

[14] Samanta, B., Gear fault diagnosis using artificial neural networks and support vector mechanics with genetic algorithms, Mechanical Systems and Signal Processing, 18(3), 625-649, 2004.

[15] Shin, H.J,, Eom, D. H., and Kim, S. S., One class support vector machines – an application in fault detection and classification, Computer and Industrial Engineering, Vol. 48, pp. 395 – 408, 2005.

[16] Staszewski, W. J., Worden. K., and Tomlinson, G. R., Time-frequency analysis in gearbox fault detection using the Wigner-ville distribution and pattern recognition, *Mechanical Systems and Signal Processing*, Vol. 11(5), pp. 673-692, 1997.

[17] Sugumaran, V., Muralidharan, V., and Ramachandran, K. I., Feature selection using Decision Tree and classification through Proximal Support Vector Machine for fault diagnostics of roller bearing, *Mechanical Systems and Signal Processing,* Vol. 21(2), pp. 930-942, 2007.

[18] Sugumaran, V., Deepak, J., Amarnath, M., and Hemanthakumar, Fault Diagnosis of Helical Gear Box using Decision Tree through Vibration

Signals, *International Journal of Performability Engineering*, Vol. 9(2), pp. 195- 208, 2013.

[19] Vyas, N. S., and Satishkumar, D., Artificial neural network design for fault identification in a rotor-bearing system, *Mechanism and Machine Theory*, Vol. 36, pp. 157-175, 2001.

[20] Weinberger, K. Q.; Blitzer J. C., Saul L. K., Distance Metric Learning for Large Margin Nearest Neighbor Classification, *Advances in Neural Information Processing Systems, Vol.* 18, pp. 1473–1480, 2006.

[21] Wuxing, L., Tse, P. W., Guiicai, Z., and Shitielin, Classification of gear faults using cumulant and the radial basis function, *Mechanical Systems and Signal Processing,* Vol. 18, pp. 381-389, 2004.

[22] Yang, M., Stronachand, A. F., McConnel, P., Third order spectral technique for the diagnosis of motor bearing conditions using artificial neural network, *Mechanical Systems and Signal Processing*, Vol. 16(2-3), pp. 391-411, 2002.

[23] Yang, B. S., Hwang, W.W., Kim, D. J., and Tan, A. C., Condition classification of small reciprocating compressor for refrigerators using artificial neural networks and support vector machines, *Mechanical Systems and Signal Processing*, Vol. 19, pp.371-390, 2005.

[24] Yesilyurt, I., Fengshou and Ball, A. D.,Gear tooth stiffness measurement using modal analysis and its use in wear fault severity assessment of spur gears, *NDT&E International*, Vol. 36, pp. 357-372, 2003.

# 8. Appendix

**A.1. Pseudo code for the modified LMNN algorithm:-**

The following is the pseudo code for the LMNN algorithm which is used in this paper. Certain changes were made from the original paper by Weinberger *et al.* [20] in order to make it better suited for fault diagnosis.

**A.1.1. Training phase algorithm**: - Initialize pseudo metric, which will be used by the Test phase algorithm

User defined input – μ (gradient step size), Training set of instances {$\vec{x}_i$}, k (number of nearest neighbors), a (gradient matrix weight)

Output – Matrix M (optimal matrix for pseudo metric)

1: $\boldsymbol{M_0} := \boldsymbol{I}$  {Initialize with the identity matrix}
2: $\boldsymbol{N} := 0$  {Initialize neighborhood matrix as 0 matrix}
3: $\boldsymbol{A_0} := \{\}$  {Initialize empty active set}
4: $\boldsymbol{G_0} := (1-\mu)\sum_{i,j \to i} C_{ij}$  {Initialize gradient}
5: $\boldsymbol{t} := 0$  {Initialize integer counter variable}
6: $\boldsymbol{i,j,l} := 0$  {Initialize integer counter variables}
7: **while** (not converged) **do**
8:    **while**($i$< number of training instances) **do**
9:       **while**($j < k$)    {k = number of nearest neighbors} **do**
10:         **if**($j \to i$ ) {if $\vec{x}_j$ if a target neighbor of $\vec{x}_i$} *do*
11:            $N[i][j] = 1$
12:         **end if**
13:      **end while**
14:   **end while**
15:   **for**∀ $(i,j)\ s.t\ N[i][j] = 1$**do**
16:      **while**($l$< K) **do**

17:                    **if** $(l \not\to i)$ {if $\vec{x}_j$ if a target neighbor of $\vec{x}_i$} **do**

18:                         $A_t = A_t \cup (i,j,l)$

19:                    **end if**

20:              **end while**

21:        **end for**

22:        $G_t = \sum_{j \to i} C_{ij} + \mu \sum_{(i, j, l) \in A_t}(C_{ij} - C_{il})$

23:        $M_{t+1} := P_S(M_t - aG_{t+1})$         {Take gradient step and project onto SDP cone}

24:        $t := t + 1$

25:        $A_{t+1} = \{\}$    {Reset the active set to empty for next iteration}

26:        $N := 0$    {Reset neighborhood matrix as 0 matrix}

26: **end while**

27: Output $M_t$

**A.1.2. Test phase algorithm: -** Classify test cases and generate accuracy and confusion matrix statistics

User defined input – Test set of instances $\{\vec{x}_i\}$, optimal matrix (M), k (number of nearest neighbors)

Output – Percentage accuracy of classification, confusion matrix of classification.

1: $M :=$ Output of training phase algorithm

2: $CActual := Get\_Class$    {Get the actual class of the test instances}

3: $CPredicted = \{\}$    {Initialize the set of predicted classes as empty set}

4: **W** = 0 {Initialize set of weights = 0}

5: **for** each instance $\vec{x}_i$ in test set $\{\vec{x}_i\}$ **do**

6:     **K-nearest** = generate k nearest neighbors from training set    {training set = total observations – test set)

7:     **for each** $\vec{x}_j$ in **K-nearest do**

8:         $C_j = class(\vec{x}_j)$

9:         $W(C_j) = W(C_j) + 1/ d(\vec{x}_i, \vec{x}_j)$

10:    **end for**

11:     **CPredicted**[$i$] = $arg - max(W)$     {Choose the index of W with highest value as the class label}

12:     **W** = 0     {Re initialize set of weights = 0}

13: **end for**

14: **Acc** = 0     {declare accuracy variable = 0}

15: $i$ = 0     {set counter = 0}

16: **while**($i$ < $test\ set\ size$) **do**

17:     if (**CActual**[$i$] == **CPredicted**[$i$]) do

18:         **Acc** = **Acc** + 1

19:     end if

20: end while

21: **Acc** = **Acc** / ($test\ set\ size$)

22: Output **Acc**

23: Output ($Get\_confusion\_matrix$(**CPredicted**, **CActual**))

The explanation of the pseudo code is as the following:-

**A.1.4. Training phase algorithm**: -The input to the training phase program is the gradient step size ($\mu$) the set of training instances {$\vec{x}_i$}, the number of nearest neighbors (k) and the gradient matrix weight ($a$) In this case, $\mu$ = 0.1 was chosen as the default value for the gradient step size. The training set and k were varied and chosen using random sub sampling method. Firstly an identity matrix of dimensions n * n is initialized where n = Number of features chosen via feature subset selection (in this study n = 4 for sound signals and n = 3, for vibration signals respectively). Initialize a Neighborhood matrix N with dimensions same as the optimal matrix and set all elements as 0. The neighborhood matrix is a bitmap which has the $i^{th}$ row and $j^{th}$ column element = 1 if $j \to i$ {$\vec{x}_j$ is a target neighbor of instance $\vec{x}_i$}. A set called Active set (A) is defined which is initially set as an empty set. The active set is a set of 3- tuples($i, j, l$) where $j \to i$ and $l \not\to i${$\vec{x}_l$ is not a target neighbor of instance $\vec{x}_i$ }. This ensures that the slack variable $\varepsilon_{i,j,l}$ is always > 0 in the set N

Now initialize the gradient matrix according to line (4) of the algorithm. The dimensions of the gradient matrix are same as the initial optimal matrix. The matrix $C_{ij}$ is defined in equation (14). An integer counter ($t$) is declared which keeps of track of number of iterations completed by the algorithm's main while loop (lines 7 to 26). Similarly 3 integer counters ($i, j, l$) are also defined which is used to generate and index the active set. Lines 8 to 14 are used to generate the neighborhood matrix for each iteration. Since only k nearest neighbors are to be checked, for each instance $\vec{x}_i$ only k immediate neighbors are checked (line 8). Furthermore it is checked whether $\vec{x}_j$ is a target neighbor of instance $\vec{x}_i$ (line 10). Accordingly the neighborhood matrix is updated. Lines 15 to 21 are used to generate the Active set. For this all ($i, j$) are chosen for which the $i^{th}$ row and $j^{th}$ column element of the neighborhood matrix is 1 (line 16). All of the k - nearest neighbors of instance $\vec{x}_i$ are checked and tested if $l \nrightarrow i$ (line 17). If yes then the 3-tuple ($i, j, l$) is added to active set N (line 18). An alternate way of updating the active set is to check whether a set of ($\vec{x}_i, \vec{x}_j, \vec{x}_l$) leads to a strictly positive value of equation (13). If it does, then the 3-tuple ($i, j, l$) is added to active set A. Lines 7 to 26 is the main loop of the training phase algorithm which generated the active set and updates the optimal matrix M using it. In this study, the convergence criteria for the main while loop (line 7) was chosen to be ten iterations of the loop. The gradient matrix is updated according to equation (18) (line 22) and the optimal matrix is updated via equation (19) (line 23). The variable $a$ in line 23 is the gradient matrix weight and is set to 0.01 as a default value in this study. The function $P_S$ is projection onto semi definite cones for the matrix M. This ensures that the optimal matrix is always positive semi-definite. In order to evaluate this function, the eigen-decomposition of the optimal matrix is needed. Its procedure is given in the last paragraph of section 4.2. Lines 25 and 26 are used to reset the neighborhood matrix and the active set to empty for the next iteration. The output is the final optimal matrix after the while loop.

**A.1.5. Test phase algorithm: -** The test phase algorithm runs once the training phase algorithm is completed. Inputs to the algorithm are the optimal matrix (M) computed in the training phase algorithm, and the set of test instances $\{\vec{x}_i\}$ and the number of nearest neighbors (k). First the Actual class labels are acquired from the test set in order to test accuracy (line 2) and put into an array called C Actual. The size of C Actual = size of test set. Now initialize an empty array called C Predicted with same size as C Actual. In order to generate the C Predicted array, a weights array (W) is initialized with value 0. The size of the array = Number of possible class labels. Lines 5 to 13 generate the C Predicted array. For each test instance $\vec{x}_i$ in the test set $\{\vec{x}_i\}$, the k-nearest neighbors of $\{\vec{x}_i\}$ is added to an array called K-nearest (line 6). In order to achieve that, the training set (total observations – test set) can be sorted based on distance from the test instance using any sorting algorithm and the first k instances from the sorted array can be picked. The class label of each of the nearest neighbors is determined (line 8) and is stored in a variable $C_j$. Now the $C_j^{th}$ index of W is updated according to equation (22) which uses equation (6) as a distance measure(line 9). Then the predicted class is classified according to a majority voting scheme based on the W array (line 11). After the classes are labelled for each test instance, the accuracy is determined based on number of matches between C Actual and C Predicted (lines 14 to 20). The confusion matrix is also generated (line 23). The details of the confusion matrix are given in section 5.